\documentclass[runningheads]{llncs}
\usepackage[hyphens]{url}
\usepackage{graphicx}
\usepackage{subcaption}
\begin{document}
\title{GETT-QA: Graph Embedding based T2T Transformer for Knowledge Graph Question Answering}
\author{Debayan Banerjee\inst{1} \and
Pranav Ajit Nair\inst{2} \and
Ricardo Usbeck\inst{1} \and 
Chris Biemann\inst{1}
}
%
%\author{Anonymous Authors}
\institute{Universität Hamburg,
Hamburg, Germany
\email{\{firstname.lastname\}@uni-hamburg.de}\and
Indian Institute of Technology (BHU),
Varanasi, India
\email{pranavajitnair.cse18@itbhu.ac.in}}
\maketitle              % typeset the header of the contribution
\begin{abstract}
In this work, we present an end-to-end  Knowledge Graph Question Answering (KGQA) system named GETT-QA. GETT-QA uses T5, a popular text-to-text pre-trained language model. The model takes a question in natural language as input and produces a simpler form of the intended SPARQL query. In the simpler form, the model does not directly produce entity and relation IDs. Instead, it produces corresponding entity and relation labels. The labels are grounded to KG entity and relation IDs in a subsequent step. 
To further improve the results, we instruct the model to produce a truncated version of the KG embedding for each entity. The truncated KG embedding enables a finer search for disambiguation purposes. We find that T5 is able to learn the truncated KG embeddings without any change of loss function, improving KGQA performance. As a result, we report strong results for LC-QuAD 2.0 and SimpleQuestions-Wikidata datasets on end-to-end KGQA over Wikidata.

%\keywords{First keyword  \and Second keyword \and Another keyword.}
\end{abstract}

\section{Introduction} \label{intro}

A Knowledge Graph (KG) is an information store where data is stored in the form of node-edge-node triples. Nodes represent entities and edges represent relationships between these entities. The aim of KGQA \cite{ijcai2021-611} is to produce answers from this KG given an input question in natural language, e.g., \texttt{Who is the father of Barack Obama ?}. Usually, the first step in KGQA is to perform Entity Linking (EL) where mention spans, e.g., \texttt{Barack Obama} representing the name of a person, place, etc., are linked to a KG node. The subsequent step is Relation Linking (RL), where the relationship of the entity to the potential answer in the KG is extracted, e.g., \texttt{father of}. 
Some KGQA systems attempt to fetch the answer based on the results of just the two steps above, which typically ends up being another entity (node) in the graph. However, for more complex questions, such as count queries or min/max aggregate queries (e.g.: \texttt{How many rivers are there in India?}) the answer does not lie in a node or edge in the graph, but instead, a formal query must be generated as a final step. To this end, semantic parsing is relevant to the problem of KGQA. Thus, our focus in this work is to generate a final SPARQL query that can be executed on the KG. 

SPARQL is a popular graph query language for querying KGs. A sample SPARQL query for the running example over the Wikidata KG looks like the following:
\\
\begin{verbatim}
  SELECT ?o WHERE { wd:Q76 wdt:P22 ?o }
\end{verbatim}

In the query above, \texttt{wd:Q76} stands for Barack Obama, while \texttt{wdt:P22} stands for the relation \texttt{father}. The \texttt{?o} variable represents the answer from the KG.

Recent works employ text-to-text (T2T) pre-trained language models (PLMs) for generating logical queries, e.g. SPARQL, from natural language questions. If the correct entity and relation IDs are already specified in the input, the accuracy of T2T models is high \cite{banerjee}. However, the absence of linked entity and relation IDs in the input presents a significant challenge to such models. PLMs are adept at generating linguistic tokens from within their weights. Yet, it is an entirely different proposition to query the KG and ground the entity and relations to specific IDs, as the variability of language creates impressive richness at generation while at the same time hampers the alignment to pre-defined KG items. 

In this work, we demonstrate a novel method by which a T2T PLM, namely T5 \cite{t5}, not only generates SPARQL queries, but also generates truncated KG embeddings, which aid in the subsequent process of grounding entities to the correct node in the KG. Our method produces strong results for end-to-end Question Answering on the LC-QuAD 2.0 and SimpleQuestions-Wikidata datasets over Wikidata KG. All code and data will be made available \footnote{\url{https://github.com/debayan/gett-qa}}.\\ \\
%All code and data used in this work can be found in the anonymised link\footnote{\url{https://drive.google.com/file/d/1kqN7rsN9fNtYy8iPxot9DLPBxrDl92Ok/view?usp=sharing}}.

%All code and data used in the publication of results in this work shall be made public upon acceptance.

\section{Related Work}
\label{relatedwork}
Early KGQA systems could be divided on the basis of whether they can handle simple \cite{info12070271} or complex questions \cite{ijcai2021-611}. In a simple question, a node-edge-node triple is a sole basis on which a question is formed, whereas in a complex question there may be more than one such triple involved. Moreover, certain KGQA systems are built specifically to handle a certain class of questions better, e.g. temporal questions \cite{jiatemporal}. 

Another way of categorising KGQA systems is whether they form a formal query \cite{10.1145/3178876.3186004,bhutani-2019-learning,qanswer,shen-etal-2019-multi,platypus} versus whether they use graph search based methods without producing an explicit query \cite{10.1145/3357384.3358016,conf/cikm/VakulenkoGPRC19,10.1145/3289600.3290956,saxena-etal-2020-improving,pullnet,graftnet,stag,sygma}.

Some KGQA systems work in a hybrid mode and can query from both KG and text-based sources. PullNet \cite{pullnet} and Graftnet \cite{graftnet} both use Relational-Graph Convolution Networks \cite{Schlichtkrull:2018jv} to handle complex questions. UNIK-QA \cite{unikqa} verbalises all structured input from KG, tables and lists into sentences and adds them to a text corpora and proceeds to perform QA over this augmented text corpora using deep passage retrieval techniques. UNIQORN \cite{uniqorn} builds a context graph on-the-fly and retrieves question relevant pieces of evidence from KG and text corpora, using fine-tuned BERT models. They use group Steiner trees to identify the best answer in the context graph. We use the results of the KG components of these hybrid systems in our evaluation in Table \ref{table:results}, as reported by UNIQORN.

Platypus \cite{platypus} and QAnswer \cite{qanswer} are two recent KGQA systems that work on Wikidata. Both of them use templates and ranking mechanisms to find the best query. 
%We include their results in our evaluation Table \ref{table:results}, as reported by \citet{uniqorn}. 
We make no use of templates in our method since this inherently limits the flexibility of a system on unseen templates.

ElneuQA-ConvS2S \cite{elneuqa} operates in a similar fashion to us, where they use a Neural Machine Translation (NMT) model to generate SPARQL queries with empty placeholders, while an entity linking and sequence labeling model fills the slots. In our case we also make use of NMT capabilities of T5 to generate a skeleton SPARQL query, however, we do not generate empty slots, and instead, generate entity and relation labels to be grounded later. \\
%We include this system in the evaluation section. \\
For simple questions, KEQA\cite{keqa} targets at
jointly recovering the question’s head entity, predicate, and tail
entity representations in the KG embedding spaces and then forming a query to retrieve the answer from a KG. Text2Graph\cite{text2graph} uses KEQA as a base, and improves on the embedding learning model by utilising CP tensor decomposition \cite{https://doi.org/10.1002/sapm192761164}. We include both these systems in our evaluation Table \ref{table:simplequestions}.\\
SGPT \cite{sgpt} and STAG \cite{stag} both use generative methods for forming the query using pre-trained language models, which is similar to what we do, however, neither of them generate the entity or relation label, or the embeddings. Instead STAG uses an external entity linking step, while SGPT attempts to generate entity and relation IDs directly. However such a method does not work well because for a KG like Wikidata, the IDs do not follow a hierarchical pattern, and hence the model is not able to predict an ID that it has not seen in training earlier. 

One of our key ideas is to enable a PLM to learn KG Embeddings. There have been some recent efforts in the same direction such as KEPLER \cite{kepler}, K-BERT \cite{kbert}, KI-BERT \cite{DBLP:journals/corr/abs-2104-08145}, CoLAKE \cite{DBLP:journals/corr/abs-2010-00309}, BERT-MK \cite{he-etal-2020-bert} and JAKET \cite{jaket}. These systems either try to inject KG embeddings into the input or intermediate layers of the model, or they try to augment the text corpora by including verbalised forms of the triple structural information. On the other hand, we ask the model to print the embeddings as output. This is a fundamentally different approach from what has been tried so far. 

A related, yet different class of systems is that of conversational QA. \\LASAGNE \cite{lasagne} and CARTON \cite{carton} are two notable systems in this category. They evaluate on the CSQA dataset \cite{csqa}, which is a conversational dataset answerable over a KG. In our case, we address only single sentence-long questions. The conversations in CSQA are arranged in sequence of turns of questions and answers. For the semantic parsing of logical forms, both LASAGNE and CARTON use a pre-defined grammar, while our approach is free of templated grammar rules. Both LASAGNE and CARTON use a Transformer architecture to generate base logical forms, however, LASAGNE uses, a Graph Attention Network \cite{velikovi2017graph} to resolve entities and relations while CARTON uses stacked pointer networks.

\begin{figure*}[htb!]
  \centering
  \begin{minipage}[b]{1.0\textwidth}
    \centering
    \includegraphics[width=1.0\textwidth]{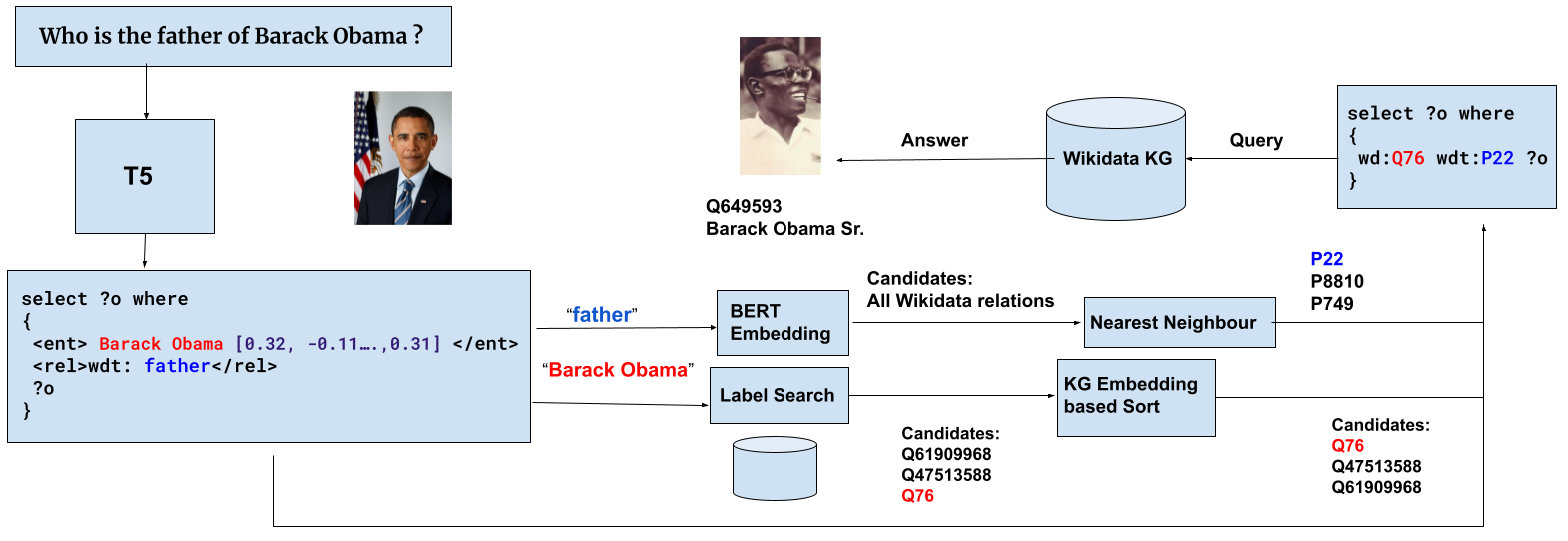}
  \end{minipage}
  \hfill
  \caption{Architecture of GETT-QA: T5 generates a skeleton SPARQL query with entity and relation labels and a truncated KG embedding string. The entity labels are grounded using label based search and embedding based sorting, while the relations are grounded using BERT embedding based sorting. The final query is executed against a KG to fetch the answer.}
\label{fig:fig1}
\vspace{-6mm}
\end{figure*}

\section{Method}

As shown in Figure \ref{fig:fig1}, our system consists of five major steps:

\begin{itemize}
    \item T5 generates a skeleton SPARQL query from input natural language question.
    \item The entity labels and truncated KG embeddings are extracted. A label search is performed to fetch entity candidates.
    \item The entity candidates are re-ranked using an embedding similarity metric.
    \item In parallel, the relation label is extracted and matched against Wikidata relations based on BERT embeddings.
    \item The final query is grounded to KG nodes and executed to receive the answer.
\end{itemize}

\subsection{Truncated KG Embeddings}

We teach T5 to generate truncated vector strings of KG embeddings. We use  TransE \cite{transe} embeddings for Wikidata entities that were provided by Pytorch-BigGraph\footnote{\url{https://github.com/facebookresearch/PyTorch-BigGraph}} \cite{pbg}. These are 200-dimensional vectors of floats. The truncated KG embeddings we use are a shorter version of the same embeddings. For most of our experiments, we use the first 10 dimensions of these embeddings, and further reduce the precision of the floats to 3 digits after the decimal. We do so since T5 is expected to produce these truncated KG embeddings while still in the text-to-text mode. In other words, T5 produces these vectors of floats considering them as a string. We use truncated KG embeddings instead of original embeddings to reduce the decoding load of T5. Our aim is not to learn the entire embedding space. Instead, we want to learn identifiers that can aid the entity disambiguation phase. We produce these truncated KG embeddings only for entities, not for relations.

\subsection{Intuition}

The initial idea behind our approach is to allow T5 to use its significant linguistic capability derived from pre-training over a large corpus for the task of semantic parsing. As a result, we use T5 to produce SPARQL tokens and entity and relation labels. 

At first glance, it may appear that the production of entity labels is sufficient for grounding to KG entity nodes. However, in most KGs, several entities share the same labels. For example in Wikidata KG, the entity IDs \texttt{Q76} and \texttt{Q47513588} both share the label \texttt{Barack Obama}. In reality, \texttt{Q76} represents the President while \texttt{Q47513588} is the entity ID for a painting of the President. As a result of such collision of labels, a further step called \textit{disambiguation} is required.  
 
The next idea is to not just rely on T5's linguistic abilities but also to try and teach the model how to generate identifiers for the entities, which can aid the grounding and disambiguation process in a subsequent step. One way could be to generate the Wikidata IDs directly. However, the IDs do not correspond in any hierarchical way to the underlying entities. For example, while \texttt{Q76} is \texttt{Barack Obama}, \texttt{Q77} is \texttt{Uruguay}. Although the IDs are close to each other, the categories are completely different. Models cannot be expected to produce accurate IDs of this nature, especially on unseen input questions. As a result, we consider other schemes of entity identifiers that exhibit some hierarchical properties.

It turns out KG embeddings fulfill these requirements handsomely, and hence we decide to use truncated KG embeddings as the "soft" identifier of our choice. Another possibility would be to generate entity descriptions instead of truncated KG embeddings, however, roughly 50\% of entities in Wikidata do not have corresponding descriptions (eg: Q67395985\footnote{\url{https://www.wikidata.org/wiki/Q67395985}}), hence we focus on generating truncated KG embeddings instead.

While the production of such truncated embeddings may also aid the grounding of relations, we do not attempt this, since Wikidata only contains a few thousand relations, while the number of entities run into several millions. For the grounding of relations we use simpler text embedding based methods, as described later in Section \ref{relationcandidates}.

\subsection{Models}

T5 \cite{t5}, or text-to-text transfer transformer, is a transformer \cite{transformers}  based encoder-decoder architecture  that uses a text-to-text approach. Every task, including translation, question answering, and classification, is cast as feeding the model text as input and training it to generate some target text. This allows using the same model, loss function, hyper-parameters, etc., across a diverse set of tasks. We experiment with fine-tuning two variants of the T5 model: \texttt{T5-Small} with 60 million parameters and \texttt{T5-Base} with 220 million parameters. For the GETT-QA system results reported in Tables \ref{table:results} and \ref{table:simplequestions} we use the \texttt{T5-Base} based model, whereas in the analysis Section \ref{snippetanalysis} we present a comparative study against \texttt{T5-Small}.

\subsection{Skeleton SPARQL}

As shown in Figure \ref{fig:fig1}, the first step of our KGQA system is to generate a skeleton SPARQL query from the given natural language question. The skeleton query consists of SPARQL tokens such as \texttt{SELECT, WHERE, \{\}}, entity and relation labels, and truncated KG embeddings, which are an array of \texttt{float}s. Some additional tokens are used to surround the entity and relation labels, such as \texttt{<ent>, </ent>,<rel>,</rel>} so that in a later step their extraction can be performed using regular-expression operations. The extraction of the labels and the truncated KG embedding  are essential for the subsequent grounding step. Notably, entity and relation IDs are not a part of a skeleton SPARQL query. 

During training via fine-tuning, pairs of questions and skeleton SPARQL queries are presented to T5. For this purpose, we pre-process the original dataset, which contains gold SPARQL queries for each question. The SPARQL query is converted to a skeleton SPARQL query by replacing the entity and relation IDs with their gold labels while appending the entity labels with a truncated KG embedding. Hence, the following gold SPARQL query:

\begin{verbatim}
select ?o where {  wd:Q76 wdt:P22 ?o }
\end{verbatim}

is converted to:

\begin{verbatim}
select ?o where 
{
 <ent>Barack Obama [...] </ent> 
 <rel>father</rel> 
 ?o
}  
\end{verbatim}

for the purposes of training T5, where the truncated KG embedding is represented by \texttt{[...]} .

\subsection{Entity Candidates} \label{entitycandidates}
During inference, when T5 generates a skeleton query, all entity and relation labels, as well as truncated KG embeddings, are extracted using regular expressions. For the entity labels, a BM-25-based \cite{Robertson1994SomeSE} label search is performed on a database of all Wikidata entity labels, out of which top-\texttt{k} candidates are retrieved per entity label. For this text search we use the Elasticsearch database\footnote{\url{https://www.elastic.co/}}. For our experiments, we fix \texttt{k} at 100.

\subsection{Entity Candidates Re-ranking and Ordering} \label{entityreranking}

The top-3 entity candidates based on label matching are retained. For the next 3 candidates, we resort to truncated KG embedding-based sorting. For each item in the list of 100 entity candidates fetched, we also fetch their gold KG embeddings, and convert them to truncated KG embeddings. For the truncated KG embedding generated by T5, we compute its dot product against the gold truncated KG embeddings fetched and re-rank them in descending order. The dot product is used as a comparator because this was the same function that was used during the production of the TransE embeddings. From this re-ranked list based on truncated KG embedding similarity, top-3 candidates are retained. 

We append these \texttt{top-3} truncated KG embedding-sorted candidates to the \texttt{top-3} label-sorted candidates, and proceed to the subsequent steps with a list of 6 candidate entities for each entity label.

\subsection{Relation Candidates} \label{relationcandidates}

We generate no truncated KG embeddings for relation IDs, as their numbers are orders of magnitudes smaller when compared to entities in Wikidata. From the relation labels generated by T5, we compute their BERT embeddings and compute the cosine similarity against the BERT embeddings of all Wikidata properties. The list of properties is sorted based on this similarity score, and the top-3 matches are considered for the subsequent steps.

\subsection{Candidate Combinations} \label{candidatepernutations}
For a generated query, each entity label and each relation label has 6 and 3 candidates each, respectively. We preserve the serial order of the entities and relations as produced in the query, and generate all possible combinations of the entities and relations, which generates several queries of the same structure but different entity and relation IDs. For example, if the query contains just one entity and one relation, the number of possible SPARQL queries generated would be $ 6 \times 3 = 18$. We execute each query on the KG in sorted order of entity and relation IDs received in previous steps. We stop when the KG returns a non-empty response. This response is considered the output of our KGQA system. We consider the top 3 beams produced by T5 decoder as probable queries. The first beam producing a valid response from the KG is considered the output of our KGQA system.

\section{Dataset}
We evaluate our approach on the LC-QuAD 2.0 \cite{lcq2} dataset, which consists of approximately 30,000 questions based on the Wikidata KG. Each question contains the corresponding SPARQL query as gold annotation. The dataset consists of a wide variety of questions, such as simple, complex, multi-hop, count, min/max, dual and boolean types. This dataset also uses the recently introduced hyper-relational \cite{StarE} structure of the Wikidata KG. \\
Additionally, we evaluate our approach on the SimpleQuestions-Wikidata \cite{wikidata-benchmark}\footnote{\url{https://github.com/askplatypus/wikidata-simplequestions}} dataset, which consists of 34,374 train questions and 9,961 test questions. This dataset is derived from the original SimpleQuestions dataset \cite{simpleqs}, which was later aligned with the Wikidata KG. A sample question from each dataset can be seen in Tables \ref{lcq2samples} and \ref{simpleqssamples}. 

\begin{table*}
  \centering
  
  \begin{tabular}{|p{4cm}|p{8cm}|}
   \hline
     Question & SPARQL      \\
    
     \hline 
    Tell me the female beauty pageant that operates in all countries and contains the word model in it's name?
                &\begin{verbatim}
SELECT DISTINCT ?sbj ?sbj_label 
  WHERE { 
  ?sbj wdt:P31 wd:Q58863414 . 
  ?sbj wdt:P2541 wd:Q62900839 . 
  ?sbj rdfs:label ?sbj_label . 
  FILTER(CONTAINS(lcase(?sbj_label), "model")) . 
  FILTER (lang(?sbj_label) = "en") 
  } 
  LIMIT 25\end{verbatim}
                   \\
        \hline
 
    \hline
  \end{tabular}
  \caption{Sample question from LC-QuAD 2.0 }
  \label{lcq2samples}
  \vspace{-10mm}
\end{table*}

\begin{table*}
  \centering
  
  \begin{tabular}{|p{4cm}|p{8cm}|}
   \hline
     Question & SPARQL      \\
    \hline
    What type of music does David Ruffin play ?
                         & \texttt{SELECT ?x WHERE \{ 
                         wd:Q1176417 wdt:P136 ?x \}
                         } \\
    % Where was Johannes Messenius born ?
    %                          & \texttt{SELECT ?x WHERE \{ wd:Q1351994 wdt:P19 ?x \} } \\
    % What kind of music is Brutal Juice ?                         & \texttt{SELECT ?x WHERE \{ wd:Q4979845 wdt:P136 ?x \}}  \\
    
    \hline
  \end{tabular}
  \caption{Sample question from SimpleQuestions-Wikidata}
  \label{simpleqssamples}
\end{table*}

%A few samples from the datasets are provided in the Appendix in Tables \ref{lcq2samples} and \ref{simpleqssamples}.

\vspace{-8mm}
\section{Evaluation}
\label{evaluation}

In Table \ref{table:results}, the results for UNIQORN, QAnswer, UNIK-QA, Pullnet, and Platypus are taken from UNIQORN \cite{uniqorn}. UNIQORN uses a test split of 4,921 questions from the original LC-QuAD 2.0 test set of 6,046 questions for all the systems. We evaluate our approach on the same split as UNIQORN. Despite our best efforts we were unable to acquire the precise KG snapshot that UNIQORN used for evaluation. UNIQORN used a Wikidata dump dated 20 April 2021, which is no longer available either in the official Wikidata repository\footnote{\url{https://dumps.wikimedia.org/wikidatawiki/entities/}}, or with the authors of UNIQORN. As a result, we ran the 4,921 questions against the NLIWOD\footnote{\url{https://www.nliwod.org/challenge}}  Wikidata dump\footnote{\url{https://hub.docker.com/r/debayanin/hdt-query-service}},\footnote{\url{https://skynet.coypu.org/#/dataset/wikidata/query}}, which is hosted on the docker hub for easy deployment, and also hosted as an API by the Universit\"at  Hamburg's SEMS group.

In Table \ref{table:simplequestions} for SimpleQuestions-Wikidata, results for KEQA and Text2Graph are taken from MEKER \cite{text2graph}. They evaluate both systems on a smaller split of the SimpleQuestions-Wikidata test set. This subset contains those questions which are valid on a custom Wikidata version they call Wiki4M. We were provided the KG by the authors of MEKER and we evaluated our system on the same.

In Table \ref{table:results}, we report macro Precision@1 based on UNIQORN's reporting preference. In Table \ref{table:simplequestions} we report macro F1 in line with MEKER. To compute metrics, we take the gold SPARQL and predicted SPARQL query and query the KG with both. We compare the results from the KG to compute true positives, false positives, and false negatives (TP, FP, FN).

\begin{table}
  \centering
  
  \begin{tabular}{|p{7cm}|p{2cm}|}
   \hline
       & P@1   \\
    
    \hline
     UNIK-QA                     & 0.005\\
     Pullnet                    & 0.011\\
     Platypus                 & 0.036\\
%     GraftNet              &  0.197\\
%     ConvS2S           & 0.269   &  - \\
     QAnswer                   & 0.308\\
     UNIQORN                   & 0.331\\
    \hline
    GETT-QA  without truncated embeddings          & 0.327 $\pm$ 0.002 \\
    GETT-QA (with truncated embeddings)          & \textbf{0.403} $\pm$ 0.0 \\
    \hline
  \end{tabular}
  \caption{Results on LC-QuAD 2.0}
  \label{table:results}
  \vspace{-10mm}
\end{table}

\begin{table}
  \centering
  
  \begin{tabular}{|p{7cm}|p{2cm}|}
   \hline
       & F1    \\
    
    \hline
     KEQA             &  0.405\\
%     SYGMA            &  0.440\\
%     STaG-QA          &  0.617\\
     Text2Graph       &  0.618\\
     
    \hline
    GETT-QA without truncated embeddings & 0.752 $\pm$ 0.004\\
    GETT-QA (with truncated embeddings) & \textbf{0.761} $\pm$ 0.002\\
    \hline
  \end{tabular}
  \caption{Results on SimpleQuestions-Wikidata}
  \label{table:simplequestions}
  \vspace{-6mm}
\end{table}
\vspace{-6mm}
\section{Results}

In Table \ref{table:results}, the bottom two rows contain the results of our system in two different settings for LC-QuAD 2.0. In the first case, our KGQA system uses the top-6 entity candidates based on label match, without the use of truncated KG embeddings for re-ranking. In the second case, we keep the top-3 entity candidates based on label match and append to it the top-3 candidates based on truncated KG embedding match. This is the same setting as described in Subsection \ref{entityreranking}. The relation candidates in both cases remain top-3 as described in Subsection \ref{relationcandidates}.

The key finding in Table \ref{table:results} is that when we compare the last two rows, our system performs better with an absolute gain of approximately 8\% when truncated KG embeddings are used. 

In Table \ref{table:simplequestions}, for SimpleQuestions-Wikidata our method without truncated KG embeddgings already outperforms the nearest competitor by an absolute margin of 13\%. This demonstrates the natural ability of T5 to predict the correct entity and relation labels given a question. Since the query structure of all the questions in this dataset is the same, no challenge is posed to T5 in trying to learn the query itself. It is noteworthy however, that after the inclusion of truncated KG embedding re-ranking, the performance remains similar. An insignificant margin of absolute improvement of 0.9\% is seen. To investigate this gap when compared to the 8\% improvement of LC-QuAD 2.0, we delve further into the nature of the two datasets and run some analysis. We find that in the case of LC-QuAD 2.0, the correct entity is found in the top-1 position of the candidate list based on text match 60\% of the time, whereas in the case of SimpleQuestions-Wikidata, this number is significantly higher at 82\%. This is so because the questions in SimpleQuestions-Wikidata contain the entity names in almost the exact form as their gold entity annotations, whereas in LC-QuAD 2.0, several entity labels are modified, misspelled or shortened by the human annotators. Hence, in the case of SimpleQuestions-Wikidata, label-only matching is in most cases sufficient, whereas in LC-QuAD 2.0, truncated KG embedding-based disambiguation holds greater importance.\\
%Although our results appear to be best in Table \ref{table:results}, we do not claim to be state-of-the-art because for LC-QuAD 2.0 we were unable to test on the precise Wikidata version that UNIQORN tested on. However, to the best of our knowledge, for SimpleQuestions-Wikidata in Table \ref{table:simplequestions} we have produced state-of-the-art results.
\vspace{-6mm}
\subsection{Limitations}

Although GETT-QA performs the best in Tables \ref{table:results} and \ref{table:simplequestions}, we do not claim state-of-the-art results on the respective datasets. This is due to a variety of reasons: as mentioned in Subsection \ref{evaluation}, we could not procure the precise Wikidata KG version as the competing systems for LC-QuAD 2.0. In the case of LC-QuAD 2.0 and SimpleQuestions-Wikidata, evaluation was performed on a truncated subset of the original test split of this dataset. As a result we can not claim that we have the best results on the entire dataset. Additionally, we could not find the code for, or run majority of systems we evaluated against, and hence resorted to using the results as reported by them.

\subsection{Additional Evaluation}
As mentioned in Section \ref{relatedwork}, STAG \cite{stag} and ElNeuQA-ConvS2S \cite{elneuqa} are comparable generative KGQA systems. For STAG, no code, data or KG versions have been made public, while for ElNeuKGQA we were unable to run their code\footnote{\url{https://github.com/thesemanticwebhero/ElNeuKGQA}} as no instructions on how to run the code exists. For STAG on SimpleQuestions \cite{simpleqs}, on a test split of 2280 questions they report F1 61.0 while we report F1 78.1 on a larger test split of 9961 questions. Lastly, ElNeuQA-ConvS2S reports an F1 of 12.9 on WikidataQA while we report 17.8. WikidataQA is a 100 question test-subset created by the authors of ElNeuQA-ConvS2S. On LC-QuAD 2.0, they report F1 of  26.9  while we report  40.3.

\vspace{-4mm}
\section{Analysis}
\label{analysis}
\vspace{-4mm}
\subsection{Error Analysis}
\vspace{-4mm}
\label{erroranalysis}

In an attempt to find the common source of errors in LC-QuAD 2.0 we find that by far the largest cause of incorrect answers is the improper grounding of entities and relations to nodes in the KG. More than 95\% of questions where the correct entities and relations were in the top-6 and top-3 candidates respectively, the right answer was eventually produced by the KG. Unfortunately, only in 41\% of the questions, the correct entities and relations were found within the 
top-k candidates. This suggests that greater focus in the area of entity and relation linking will produce better results. One may also increase the size of \texttt{k}, at the cost of increased run time. 

Less than 1\% of the queries generated had incorrect truncated KG embedding length (e.g.: 11 instead of 10) however these were handled in code appropriately. Less than 1\% of queries generated were improper SPARQL, where a critical keyword was missing rendering the query syntactically incorrect. This suggests that T5 learns how to generate valid SPARQL queries to a large extent. This is consistent with the findings of \cite{banerjee} where T5 crosses 90\% accuracy when provided with grounded entity and relation IDs with their labels.

To explore the issue of lack of correct entities and relations in candidate lists, we observe that while 60\% of questions contain the correct entity ID in the top-100 label-based search candidates, by the time we reduce this list to top-6, only 49\% questions remain with the correct entity ID in the candidates list. On the other hand, for relation candidates, 45\%
of questions contain the correct relation IDs in the top-3 relation candidates.

When looking at the category of questions that return incorrect KG responses, irrespective of entity and relation grounding, we find that \texttt{COUNT} queries are the most common. This happens due to a quirk in SPARQL format. If a \texttt{COUNT} query is built around a family of triples that do
not exist in the KG, the KG responds with \texttt{count = 0}, instead of producing a \texttt{NULL} value or an \texttt{ERROR}. This means that the very first query to be executed on the KG with the correct \texttt{COUNT} SPARQL syntax will return a valid response, even if it is \texttt{count = 0} and we no longer explore subsequently ranked queries. \\
In the case of SimpleQuestions-Wikidata, all errors are either due to incorrect entity or relation linking. The model produced an accuracy of 70\% for entity linking and 94\% for relation linking. 

% For some questions, even incorrectly linked entities and relations may produce the right answer, which is why the F1 reported is higher than 70\%. For example, in the SimpleQuestions-Wikidata dataset, we have the following question:

% \texttt{What language is Viper in ?}

% The gold entity and relation IDs are as follows:

% \texttt{Q1189775} - Viper (action-adventure TV series)

% \texttt{P364} - original language of film or TV show

% The predicted entity and relation by the model are as follows:

% \texttt{Q7933486} - Vipers (2008 television film directed by Bill Corcoran)

% \texttt{P364} - original language of film or TV show

% Both the pairs of entity and relations when placed in a SPARQL query return the same response of \texttt{Q1860} which stands for the English language. Such a phenomenon occurs because there may be several duplicate entities in Wikidata KG.

\begin{figure}[h!]
    \centering
    \includegraphics[width=0.9\textwidth]{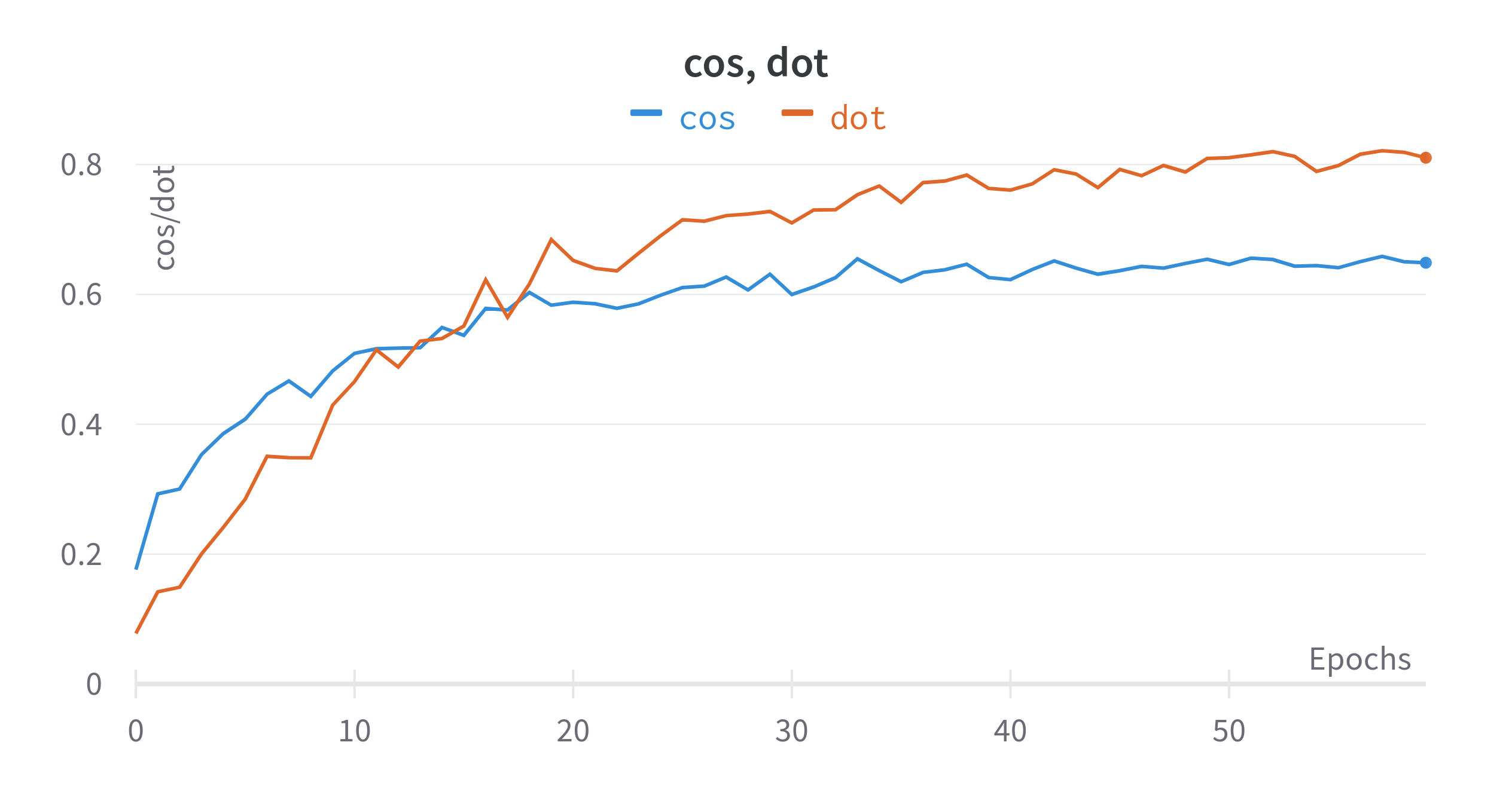}
    \caption{Cosine and Dot Product based similarities of truncated KG embeddings}
    \label{fig:cosdot1}
    \vspace{-3mm}
\end{figure}

\vspace{-4mm}
\subsection{Truncated KG embedding Learning}
\label{snippetanalysis}
We discover that T5 is able to produce a vector of floats while still in the text-to-text mode of decoding.  For this functionality, No change of loss function or decoding scheme is made in our experiments. It effectively learns a simplified embedding space, but with certain limitations. To further explore this ability of T5, we performed some additional experiments on the 200 questions dev set of LC-QuAD 2.0 with \texttt{T5-Small}.

In Figure \ref{fig:cosdot1},  we compare how the model learns the embedding space with each epoch of training. 
The TransE embeddings have an angular component and a magnitude component. Since the dot operation was used to train the original embeddings, the magnitude of each embedding may be greater than 1. It appears in Figure \ref{fig:cosdot1} that around step 20 the angular component of the embeddings has been learned by T5 to the best of its abilities, and it proceeds to learn the magnitude component (denoted by the orange line) further.

%We observe that the model quickly reaches peak performance of matches on the dev set and stagnates, while it continues to improve its knowledge of the embedding space.

% , while in Figure \ref{fig:matches1} the rate at which the snippet-less skeleton SPARQL query is learned is depicted. For this specific graph, we remove the snippet part of the skeleton SPARQL query and measure the ability of the model to generate accurately the SPARQL tokens and entity and relation labels. 
\begin{figure}[h!]
    \centering
    \includegraphics[width=0.9\textwidth]{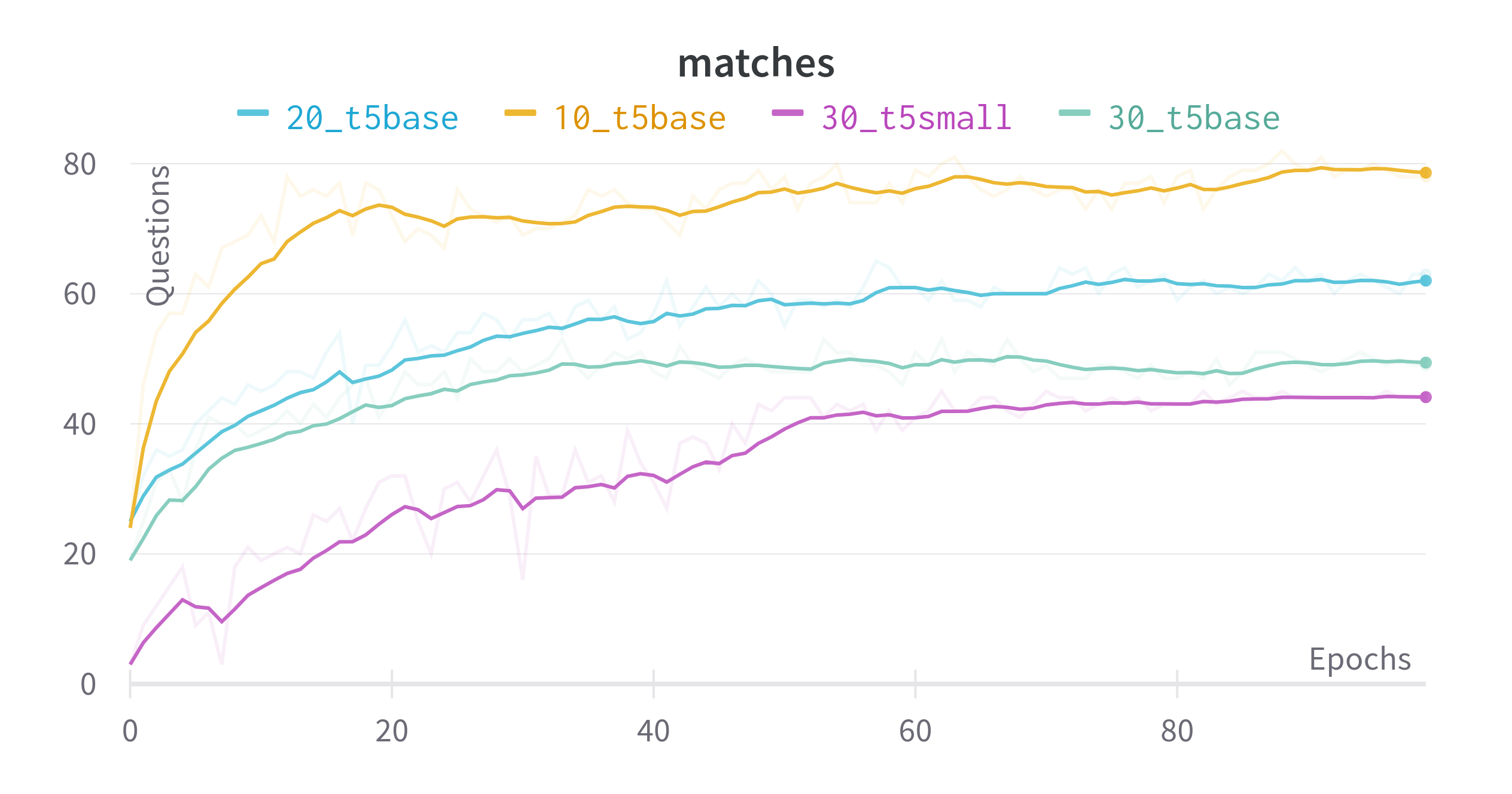}
    \vspace{-5mm}
    \caption{Dev Set matches for varying truncated KG embedding lengths}
    \label{fig:matches_3}
    \vspace{-3mm}
\end{figure}

\begin{figure}[h!]
    \centering
    \includegraphics[width=0.9\textwidth]{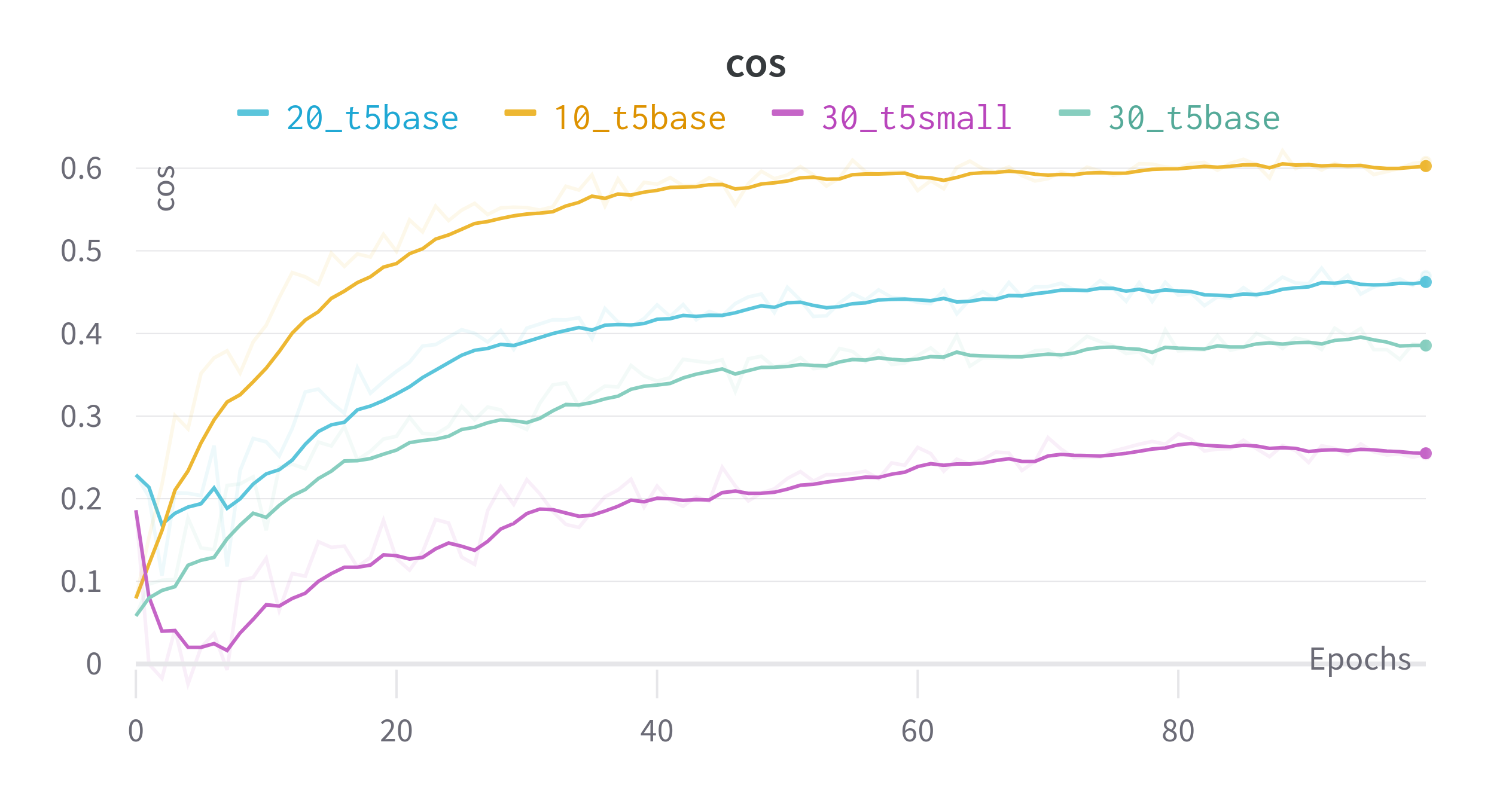}
    \vspace{-5mm}
    \caption{Cosine similarities for varying truncated KG embedding lengths}
    \label{fig:cos_3}
    \vspace{-3mm}
\end{figure}

\begin{figure}[h!]
    \centering
    \includegraphics[width=0.9\textwidth]{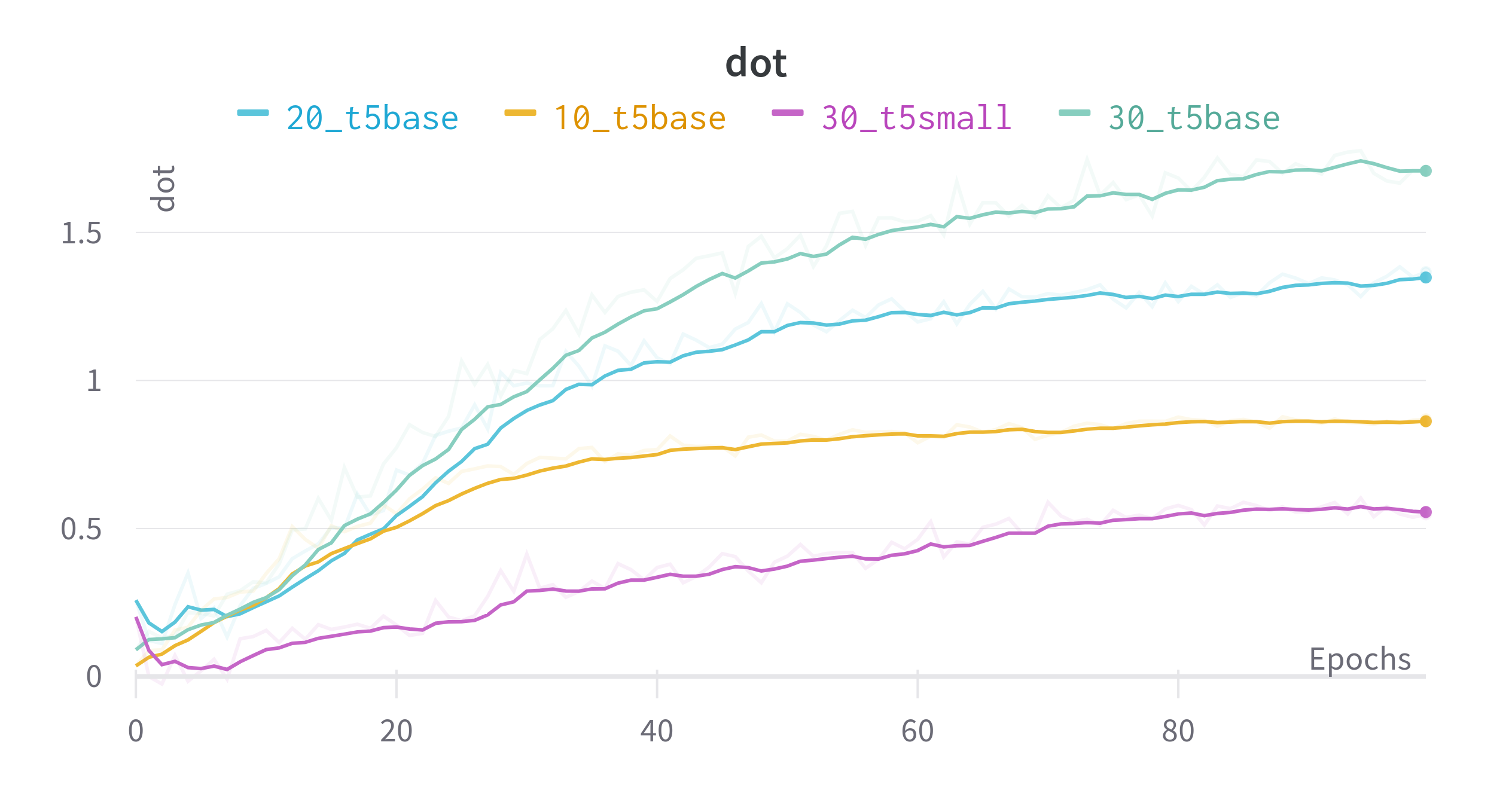}
    \vspace{-5mm}
    \caption{Dot Product for varying truncated KG embedding lengths}
    \label{fig:dot_3}
    \vspace{-3mm}
\end{figure}

 In Figures \ref{fig:matches_3}, \ref{fig:cos_3} 
and \ref{fig:dot_3}, we chart the LC-QuAD 2.0 dev set performance of \texttt{T5-Small} and \texttt{T5-Base} in varying truncated KG embedding lengths. For \texttt{T5-Small} (pink line) we set the truncated KG embedding length at 30, so this should only directly be compared against \texttt{T5-Base} with truncated KG embedding length 30 (green line). We see in Figures \ref{fig:cos_3} and \ref{fig:dot_3} that in cosine and dot product metrics \texttt{T5-Small} consistently under-performs. 

 However, in the matches metric, which only looks at the keywords and labels produced in the skeleton SPARQL query (truncated KG embeddings have been removed from the generated query), the two reach similar performance. This suggests that a larger number of parameters helps learn the embedding space better, but for the textual component the extra number of parameters of \texttt{T5-Base} remain unused.

\begin{figure}[t]
\begin{tabular}{cc}
  \includegraphics[width=60mm]{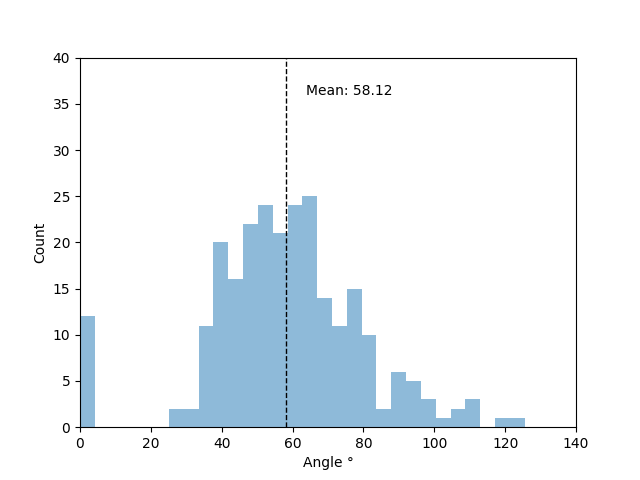} &   \includegraphics[width=60mm]{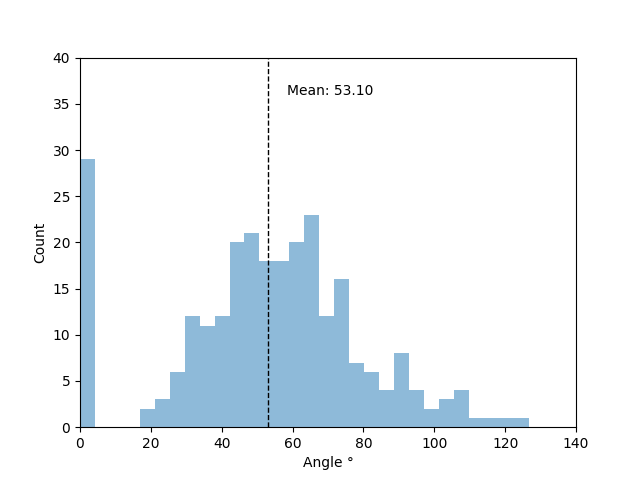} \\
(a) 10 epochs & (b) 20 epochs \\[6pt]
 \includegraphics[width=60mm]{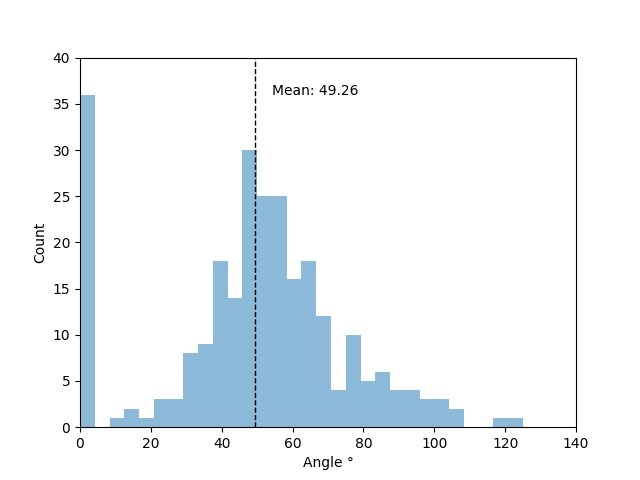} &   \includegraphics[width=60mm]{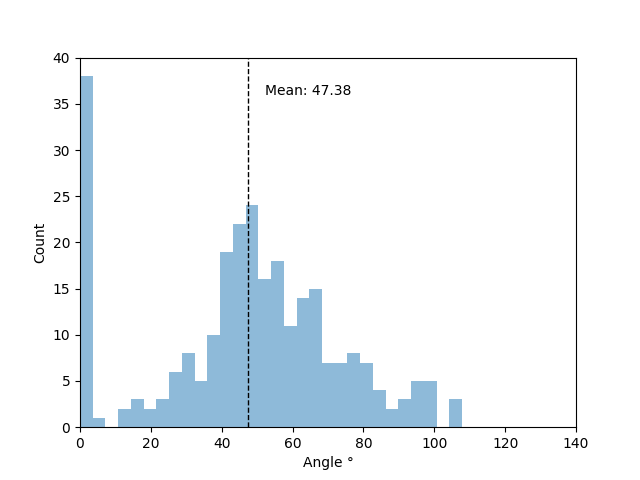} \\
(c) 30 epochs & (d) 40 epochs 
\end{tabular}
\caption{Distribution of angular difference between gold and predicted truncated KG embeddings on LC-QuAD 2.0 dev set. The mean angular difference can be seen reducing as the epochs progress, suggesting that the model is learning the embedding space.}
\label{angularplots}
\vspace{-5 mm}
\end{figure}

 In Figures \ref{fig:cos_3} and \ref{fig:dot_3}, we focus only on \texttt{T5-Base}'s performance with varying truncated KG embedding lengths. It seems larger truncated KG embedding lengths produce worse cosine similarity and label matching performance (green vs yellow), but better dot products. One explanation of the behavior could be that as we increase the truncated KG embedding lengths, the curse of dimensionality kicks in, and hence the angular space becomes less discriminable, as most vector pairs tend to differ in angle by $\pi/2$. At this point, the model begins to learn the magnitude aspect of the embedding and excels at it sooner than the models handling smaller truncated KG embedding lengths.

 With a truncated KG embedding length of 10 (yellow line), we see the best label match accuracy and hence we persist with this family of models for reporting results in Table \ref{table:results}.

 In Figure \ref{angularplots}, we plot the distribution of the angular difference between the gold and predicted truncated embeddings on the LC-QuAD 2.0 dev set. The model seems to learn the embedding space in two distinctly different manners: firstly, for several entities it is able to print the exact embedding, with an angular difference of 0$^{\circ}$. Secondly, for the entities which it is unable to learn the embedding of exactly, it produces a more familiar distribution, where the mean shifts every few epochs, reducing the angular difference. This suggests that the model is learning the embedding space effectively.

\begin{table}[h]
  \centering
  
  \begin{tabular}{|p{3cm}|p{1cm}|}
  \hline
      &   F1   \\
    
    \hline
     3 LS + 3 TS              &    0.365     \\
     3 TS + 3 LS              &    0.331    \\
     3 LS + 0 TS              &    0.289     \\
     0 LS + 3 TS              &    0.236     \\
     6 LS + 0 TS              &    0.319        \\
     0 LS + 6 TS              &    0.256        \\
    \hline
  \end{tabular}
  \caption{Effects of ordering entity candidates differently. LS = Label sorted, TS = Truncated KG embedding sorted }
  \label{table:ordering}
  \vspace{-5mm}
\end{table}

\vspace{-5mm}
\subsection{Candidate Ordering}

As mentioned in Subsection \ref{entityreranking}, our results reported in Table \ref{table:results} come from a configuration of our system where the entity candidates are layered in two parts: the first three candidates are sorted based on label match, while the bottom three candidates are sorted based on the truncated KG embedding dot product similarity. It is observed in Table \ref{table:ordering} that the ordering of these two categories affects the eventual accuracy strongly. We take 200 questions at random from the LC-QuAD 2.0 test set and perform experiments to ascertain how the ordering of candidates affects accuracy. In the first row of the table, three entity candidates based on label sorting are followed by three candidates of truncated KG embedding sorting, while the next row of the table shows the result when we keep truncated KG embedding-sorted candidates above label-sorted candidates. The results show that keeping truncated KG embedding-sorted candidates at the top reduces accuracy, and hence, label-based matching for entities remains a stronger mode of fetching correct candidates. This is no surprise, since not all labels have multiple entity candidates requiring disambiguation. Kindly note that the accuracy drops because an earlier query formed due to candidate combinations (as explained in Subsection \ref{candidatepernutations}) returns a non-empty result from the KG and this response turns out to be incorrect.

In the next two rows, we see how excluding either label-sorted candidates, or truncated KG embedding-sorted candidates entirely affects accuracy. Once more we see that label-sorted candidates still perform better when used in isolation. However, the crux of the table is the first row, i.e., when both the categories are appropriately sequenced and used in tandem, the accuracy is best.

In the bottom two rows, we see the effect of changing the number of candidates. It is no surprise that increasing \texttt{k} from 3 to 6 increases accuracy since more correct entities are included in the list. However, some systems also perform worse in such settings as the noise may increase adding to the disambiguation load. In our system, this does not seem to be the case.

Increasing the value of \textit{k} further imposes a large cost on the run time of the system affecting the user experience adversely. Since the candidate combination step has exponential complexity, which depends on the number of entities and relations in a query, we need to keep the number of candidates in check. Too many candidates will produce too many SPARQL queries and the user must wait for all of them to be executed on the KG till one of them responds validly.

In our choice of setting \texttt{k=6} for entities and \texttt{k=3} for relations, we observe that on the test set of LC-QuAD 2.0, our system has an average response time of 1.2 seconds per question, which from a user experience perspective seems like an acceptable response time.

% \section{Simple vs Complex Queries}

% We use the T5 model without any decoding constraints, so it has the ability to output an infinite range of vocabulary. This allows it to generate any kind of SPARQL query possible. There is no special handling for complex queries required for min/max/count style aggregation queries, or queries that require copying of literals from input to output \cite{banerjee}.  If the model has seen enough examples of it during training, it can generate them during inference as well. 

\vspace{-3mm}
\section{Hyperparameters and Hardware}
\vspace{-3mm}

For the evaluation of LC-QuAD 2.0 in Table \ref{table:results}, we fine-tune our models for 50 epochs with a learning rate of \texttt{1e-04} with the Adam optimizer \cite{adam}. For SimpleQuestions in Table \ref{table:simplequestions} we fine-tune for 25 epochs, roughly half of LC-QuAD 2.0, since the train set is roughly twice as large as LC-QuAD 2.0. We use a batch size of 8. During this phase we had access to \texttt{NVIDIA GeForce RTX 2080 Ti/1080 Ti} graphics cards with approximately 11GB of video memory. We do not fix a seed during training, and train and infer three times. We report mean and standard deviation for the three runs in the respective tables.

For the analysis in Section \ref{analysis}, we fine-tune our models for 100 epochs with a learning rate of \texttt{1e-05} with the Adam Optimizer \cite{adam} and we use a batch size of 20. During this phase we had access to larger GPUs, namely \texttt{NVIDIA RTX A6000} with 48GB of memory and RTX A5000 with 24GB memory. 

\vspace{-3 mm}
\section{Conclusion and Future Work}
\vspace{-3 mm}

In this work, we presented a novel KGQA pipeline called GETT-QA.
We use no external entity or relation linking tool, and still achieve strong results on LC-QuAD 2.0 and SimpleQuestions-Wikidata datasets. Additionally, we discover the ability of T5 to learn KG embeddings. We demonstrate that in certain situations this ability helps in better question answering performance.

In future work, we will explore the ability of T5 in generating similar truncated KG embedding based queries with modified loss functions and a customised architecture towards the penultimate layers of the models, so that embeddings can be generated with more standard loss functions meant specifically for learning embeddings. Additionally, suitable identifiers other than embeddings can also be explored, for example, text description based identifiers.

\section{Acknowledgements}

This research was supported by grants from NVIDIA and utilized NVIDIA 2 x RTX A5000 24GB. Furthermore, we acknowledge the financial support from the Federal Ministry for Economic Affairs and Energy of Germany in the project CoyPu (project number 01MK21007[G]) and the German Research Foundation in the project NFDI4DS (project number 460234259). This research is additonally funded by the ``Idea and Venture Fund`` research grant by Universit\"at Hamburg, which is part of the Excellence Strategy of the Federal and State Governments.

% \section{Limitations}

% As Table \ref{table:results} depicts, our system only reaches a 0.42 F1 score on LC-QuAD 2.0, leaving a large room for improvement on the given dataset. Moreover, our datasets, while containing widely varying questions, do not address the challenge of compositional generalization, which is currently a popular problem in KGQA. We limit our evaluation to Wikidata KG, because it is constantly updated, and as a research group, we consider it an important KG for research. In comparison, Freebase development has remained dormant for close to a decade, whereas  DBPedia eventually bases its releases on Wikidata KG.

\bibliographystyle{splncs04}
\bibliography{anthology}

\end{document}